\documentclass[runningheads]{llncs}
\usepackage[utf8]{inputenc}

\usepackage{amsmath}
\usepackage{amssymb}
\usepackage{siunitx}
\usepackage{caption}
\usepackage{subcaption}
\usepackage{float}
\usepackage{algorithm}
\usepackage[noend]{algpseudocode}
\usepackage{hyperref}

\usepackage{tikz}
\usepackage{tikz-3dplot}
\usetikzlibrary{positioning}

\makeatletter
\def\blfootnote{\xdef\@thefnmark{}\@footnotetext}
\makeatother

\newcommand{\bs}[1]{\boldsymbol{#1}}

\makeatletter
\define@key{circlekeys}{cx}{\def\cx{#1}}
\define@key{circlekeys}{cy}{\def\cy{#1}}
\define@key{circlekeys}{cz}{\def\cz{#1}}
\define@key{circlekeys}{cr}{\def\cr{#1}}
\define@key{circlekeys}{ca}{\def\ca{#1}}

\tikzdeclarecoordinatesystem{circlecoord}{%
	\setkeys{circlekeys}{#1}%
	\pgfpointxyz{\cx + \cr * cos(\ca)}{\cy + \cr * sin(\ca)}{\cz}}

\begin{document}

\blfootnote{28th International Conference on Artificial Neural Networks, 2019 \\ The final authenticated publication is available online at \url{https://doi.org/10.1007/978-3-030-30508-6_2}}

\title{Distortion Estimation Through Explicit Modeling of the Refractive Surface}

\author{{Sz}abolcs~P\'avel\inst{1, 2} \and {Cs}an\'ad S\'andor\inst{1, 2} \and Lehel {Cs}at\'o\inst{1}}

\institute{Faculty of Mathematics and Informatics, \\
    Babe\c{s}-Bolyai University, Kog\u{a}lniceanu 1,
	Cluj-Napoca, Romania
	\and
	Robert Bosch SRL, Some\c{s}ului 14,
	Cluj-Napoca, Romania
    \email{\{szabolcs.pavel,csanad.sandor,lehel.csato\}@cs.ubbcluj.ro}
}

\maketitle

\begin{abstract}
	Precise calibration is a must for high reliance 3D computer vision algorithms.
	A challenging case is when the camera is behind a protective glass or transparent object: due to refraction, the image is heavily distorted;
	the pinhole camera model alone can not be used and a distortion correction step is required.
	By directly modeling the geometry of the refractive media, we build the image generation process by tracing individual light rays from the camera to a target.
	Comparing the generated images to their distorted -- observed -- counterparts, we estimate the geometry parameters of the refractive surface via model inversion by employing an RBF neural network.
	We present an image collection methodology that produces data suited for finding the distortion parameters and test our algorithm on synthetic and real-world data.
	We analyze the results of the algorithm.

	\keywords{Inverse models \and Image distortions \and Calibration}

\end{abstract}

\section{Introduction}

Video-cameras are widely used in different robotic and automated driving applications. These applications frequently employ the pinhole camera model to make the association between the outside world and image pixels.
The parameters of the model are found using a camera calibration procedure: done either statically, using calibration patterns (e.g.\ checkerboards, see Fig.~\ref{fig:boards}), or with self-calibration, where the geometric constraints of the scene are leveraged.
The camera model is a first step towards image analysis \cite{szeliski2010computer}: it considers the way a 3D scene -- including geometry, lights, materials, etc. -- is mapped to a 2D image.
This complex mapping is further split into three stages: a geometric, a photo-metric and a sampling stage.
Our work considers the first stage, the geometric part: the association of 3D points with 2D pixels by explicitly modeling the refraction caused by a refractive material present between the camera and the object (e.g. protective covers made out of glass or transparent plastic materials).
To find this association, we have to follow light rays hitting a given pixel of the sensor.
Due to the refractive material, tracing is more complex: as light enters or leaves a denser media, it changes direction, resulting in deviations from the pinhole model;  called \emph{image distortions}, as shown in Fig.~\ref{fig:schematic}(a).
We construct the \emph{forward model} $f_{\theta}(\bs{p}): \Omega \rightarrow \mathbb{R}^3$, where -- knowing the camera parameters, the refractive media and scene characteristics, jointly denoted as $\theta$ -- we map a pixel $\bs{p}$ from the image $\Omega \subset \mathbb{R}^2$ to a point in the scene.
We implement this function as a raycasting algorithm -- see Sect.~\ref{sec:raycast} --, allowing us to generate images given a set of parameters.
After constructing the forward model, by using  model inversion, we fit the parameters of the refractive media to a set of observations, given as displaced points.
We build an RBF-network \cite{haykin2009neural} based parametric model of the thickness of the refractive media and use ML estimation \cite{bishop2006} to infer the optimal parameters that generated the distortions.

Our contribution in this work is three-fold: (1) we introduce a parametric model of the refractive media and derive the geometric image formation process in the presence of the refractive media (2) we provide a methodology to estimate the model parameters using a static calibration setup with checkerboard patterns (3) we present our experiments where we estimate the image distortions induced by a conic glass surface in a synthetic, as well as in a real-world scenario.

\section{Related Work}\label{sec:related}

Most 3D computer vision algorithms assume that the pinhole camera model precisely describes image optics but this is not the case in the presence of \emph{geometric} image distortions, where pixels are displaced compared to their expected positions.
Without estimating and correcting the images subject to these distortions, 3D computer vision algorithms often loose performance or simply fail to produce meaningful results.
To put our work into context, we review some methods for correcting image distortions.

Estimating image distortions is usually a building block of the \textbf{camera calibration algorithm}. These algorithms can be static or can use self-calibration.
Static calibration \cite{tsai1987versatile,zhang2000flexible} techniques use objects with previously known patterns and sizes to extract camera -- distortion -- parameters.
These algorithms provide the highest accuracy, but require the presence of a calibration object.
Self-calibration methods \cite{devernay2001straight,cefalu2016structureless,fitzgibbon2001simultaneous} use geometric constraints of the imaging system to estimate camera parameters.
These methods are less accurate then static methods, but are more flexible and can be used for online calibration. In our method we perform static calibration using a checkerboard pattern.

Specific algorithms use explicit \textbf{distortion models} of differing complexity; the most popular being Brown's polynomial distortion model \cite{brown1966decentering} for radial and tangential distortions.
The division model \cite{fitzgibbon2001simultaneous} uses an even simpler model with a single parameter to estimate radial distortions only.
Using the radial distortion model as above, one can estimate the center of the distortions \cite{hartley2007parameter}.
Fish-eye lenses create specific image distortions and the \emph{field of view} model explicitly considers those distortions \cite{devernay2001straight}, describing the FOV of an ideal fish-eye lens.
Lastly, we mention the \emph{rational function} distortion model \cite{claus2005rational}, the most similar to our model, as they lift the 2D pixels into the 3D space and associate rays to individual pixels.
Our method also traces 3D light rays, but we instead directly parameterize the refractive surfaces which generate the distortions.

The work of Agrawal et. al. \cite{agrawal2012theory} sets up distortions for images taken through flat refractive surfaces (they use a water tank for their experiments), therefore modeling explicitly the \textbf{distortions from light refraction}.
They use the theory of non-central cameras \cite{sturm2004generic} and multi-view geometry in the presence of refractive media \cite{chari2009multiple}.
Morinaka et. al. \cite{morinaka20183d} presents static camera calibration and 3D reconstruction in challenging setups like images taken through a wine glass.
They use the ``raxel'' imaging model \cite{grossberg2005raxel} with two calibration planes, with polynomial mappings between image pixels and corresponding points on the planes.
This model is closest to ours, the main difference being that in their method the refractive media is treated as a black box, while we directly model the refractive surface, giving us a global model.

\section{Refractive Surface Model}\label{sec:ref_surf}

\begin{figure}[t]
    \centering
    \begin{tabular}{@{}p{0.48\linewidth}@{}p{0.48\linewidth}@{}}
      \begin{minipage}{\linewidth}
	\begin{tikzpicture}[scale=0.6]
	\draw [fill=red] (0, 0) circle [radius=0.05] node [above left] {$\mathcal{C}$};
	\draw [thick] (0, 0) -- (1, 0.5) -- (1, -0.5) -- cycle;
	\draw [color=red, thick] (1.0, 0.20365583772877596) -- (1.0, 0.07559054806858273);
	\draw [fill=blue, fill opacity=0.3] (3.6984631039295426, -1.2898992833716565) arc (20:60:5) --
	(2.500000000000001, 3.0621778264910704) arc (60:20:7) -- cycle;
	\draw [thick] (0, 0) -- (2.83022221559489, 0.21393804843269626);
	\draw [dashed] (1.2981333293569342, -1.0716371709403822) -- (5.128355544951824, 2.142300877492314);
	\draw [thick, ->] (2.83022221559489, 0.21393804843269626) -- (3.5962666587138683, 0.8567256581192355) node [above left] {$\bs{n}_1$};
	\draw [thick] (2.83022221559489, 0.21393804843269626) -- (4.728551289842027, 1.0228969809888548);
	\draw [dashed] (2.27345787990973, -0.7012017251492257) -- (6.365280229796893, 2.172296118414242);
	\draw [->, thick] (4.728551289842027, 1.0228969809888548) -- (5.54691575981946, 1.5975965497015485) node [above] {$\bs{n}_2$};
	\draw [thick] (4.728551289842027, 1.0228969809888548) -- (7.67393816018753, 1.5628423046918132);
	\draw [color=red, thin,dashed] (0, 0) -- (7.67393816018753, 1.5628423046918132);
	\draw [thick,fill=black!50!red] (7.8, 1.57) circle (.3em) -- +(-100:3em) -- +(-80:3em) node [below left=.3em] {Object}   -- cycle;
	\node at (4.0, 2.5) {$\eta_{1}$};
	\node at (2.65, 2.5) {$\eta_{2}$};
	\draw [->, red] (0.5, 1) node [align=center, above] {Distortion \\$\Delta\bs{p}$} -- (0.99, 0.1);
	\end{tikzpicture}
      \end{minipage}
      &
      \begin{minipage}{\linewidth}
        \hfill
\tdplotsetmaincoords{70}{125}
\begin{tikzpicture}[tdplot_main_coords, scale=0.9,>=latex]
	\clip ( 6.,-.5,-.2) rectangle (-5,0,.8);

	\coordinate (O) at (0, 0, 0);
	\coordinate (TopConeC) at (0, -3, 1);
	\coordinate (BotConeC) at (0, -3, -1);

	\draw[->] (O) -- (1, 0, 0) node[above]{$x$};
	\draw[->] (O) -- (0, 1, 0) node[above]{$z$};
	\draw[->] (O) -- (0, 0, -1) node[left]{$y$};

	\draw[fill=red] (O) circle (2pt);
	\draw[thick, red, fill=red, fill opacity=0.3, shading angle=140]
			(-0.183, 0.286, 0.138) -- (0.183, 0.286, 0.138) --
		    (0.183, 0.286, -0.138) -- (-0.183, 0.286, -0.138) --
	        cycle;
	\draw[thick, color=red] (O) -- (-0.183, 0.286, 0.138);
	\draw[thick, color=red] (O) -- (0.183, 0.286, 0.138);
	\draw[thick, color=red] (O) -- (0.183, 0.286, -0.138);
	\draw[thick, color=red] (O) -- (-0.183, 0.286, -0.138);

    \draw[dashed] plot[variable=\ca, domain=25:135](circlecoord cs:cx=0, cy=-3, cz=1, cr=4.3, ca=\ca);
	\draw[dashed] plot[variable=\ca, domain=25:135](circlecoord cs:cx=0, cy=-3, cz=-1, cr=4.088, ca=\ca);
	\draw[fill] (TopConeC) circle (2pt);
	\draw[fill] (BotConeC) circle (2pt);
	%
    \draw[thick, fill=blue, fill opacity=0.2, shading angle=140]
    plot[variable=\ca, domain=75:105](circlecoord cs:cx=0, cy=-3, cz=1, cr=4.3, ca=\ca) --
    (circlecoord cs:cx=0, cy=-3, cz=1, cr=4.3, ca=105) -- (circlecoord cs:cx=0,cy=-3,cz=-1,cr=4.088,ca=105) --
    plot[variable=\ca, domain=105:75](circlecoord cs:cx=0, cy=-3, cz=-1, cr=4.088, ca=\ca) --
    (circlecoord cs:cx=0, cy=-3, cz=1, cr=4.3, ca=75) -- (circlecoord cs:cx=0,cy=-3,cz=-1,cr=4.088,ca=75);

	\draw[<->] (TopConeC) -- (BotConeC) node[midway, above, sloped] {$2$ cm};

	\draw[dashed] (O) -- (0, 0, 1);
	\draw[thin] (TopConeC) -- (circlecoord cs:cx=0, cy=-3, cz=1, cr=4.3, ca=90);
	\draw[dashed] plot[variable=\ca, domain=75:105](circlecoord cs:cx=0, cy=-3, cz=1, cr=3, ca=\ca);
	\draw[thin] (TopConeC) -- (circlecoord cs:cx=0, cy=-3, cz=1, cr=4.3, ca=75);
	\draw[thin, <->] (TopConeC) -- (circlecoord cs:cx=0, cy=-3, cz=1, cr=3, ca=105) node[midway, above, sloped, yshift=-1pt] {$3$ cm};
	\draw[thin, <->] (circlecoord cs:cx=0, cy=-3, cz=1, cr=3, ca=105) -- (circlecoord cs:cx=0, cy=-3, cz=1, cr=4.3, ca=105) node[midway, above, sloped, yshift=-1pt] {$1.3$ cm};

	\draw[thin] (BotConeC) -- (circlecoord cs:cx=0, cy=-3, cz=-1, cr=4.088, ca=75);
	\draw[thin] (BotConeC) -- (circlecoord cs:cx=0, cy=-3, cz=-1, cr=4.088, ca=105);

	\draw plot[variable=\ca, domain=75:105](circlecoord cs:cx=0, cy=-3, cz=-1, cr=1.5, ca=\ca) node[above,yshift=-.2ex,xshift=-1ex] {$30^{\circ}$};
	\draw[<->] (BotConeC) -- (circlecoord cs:cx=0, cy=-3, cz=-1, cr=4.088, ca=75) node[midway, below, sloped] {$4.1$ cm};

\end{tikzpicture}
      \end{minipage}
		\\
		\centerline{(a)} & \centerline{(b)}
	\end{tabular}
	\vspace*{-1.5em}
	  \caption{\textbf{Refraction (a):}
    	as light enters or leaves a material, it changes direction; governed by incident angles and normal vectors $\bs{n}_1$ and $\bs{n}_2$, leading to a ``shift'' $\Delta\bs{p}$ in pixel position.
      In our \textbf{experiment (b):} the camera is inside a conical glass object. The blue area denotes the $30$ degree by $2$ cm patch where we consider the uneven surface, see  Sect.~\ref{sec:ref_surf} (best viewed in color).}
    \label{fig:schematic}
\end{figure}
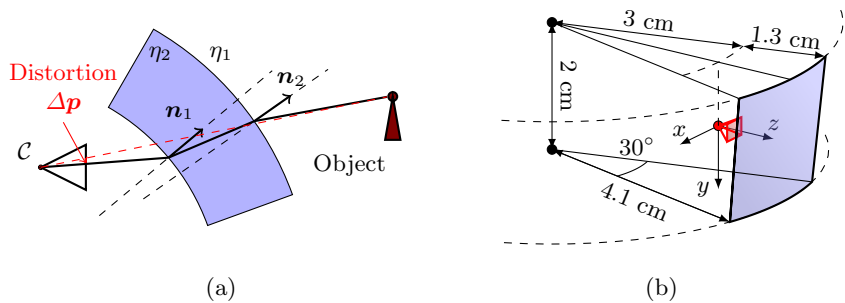

We model our refractive media as a ``thick'' cone slice, as in Fig.~\ref{fig:schematic}(a).
The inner and outer cones have the same aperture, and the centers are such that the thickness $\Delta r$ of the media is constant.
We constrain the position of the cone such that the main axis is parallel with the $y$-axis of the camera coordinate system and that the cones shrink in the positive (downwards pointing) $y$ direction, shown in Fig.~\ref{fig:schematic}(b).
We parameterize the cone surface with its height relative to the apex $s_1 \in [0, h]$ and a polar angle $s_2 \in [-\pi, \pi]$, where a polar angle of $0$ describes the points on the $YOZ$ plane, with positive $z$ values.
We use bold notation for the two-dimensional vectors: $\bs{s}=(s_1,s_2)$.

\begin{figure}[t]
	\centering
	\includegraphics[scale=1.0]{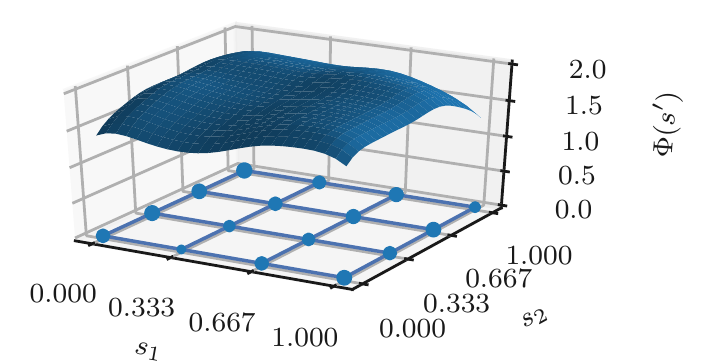}
	\caption{Sample offsets generated as a linear combination of $4\times4$ RBF kernels distributed on a regular grid. Dot sizes denote the -- positive -- kernel amplitudes.}
	\label{fig:rbfsample}
\end{figure}

To model the uneven refractive media -- implicit the distortions -- we add a parametric surface in the radial direction to the cone; this radial offset is defined as a Radial Basis Function network (RBF) \cite{park1991universal} where the inputs are the cone coordinates. 
We choose an RBF network because they are universal function approximators given a sufficient number of centers, and can be used to estimate arbitrary surfaces. In practice we show that a relatively small number of parameters is enough to model complex surfaces -- see Sect. \ref{sec:exp:synth}.
The RBF centers $\left\{\bs{s}_{ij}\right\}$, ${i,j=\overline{1, N}}$ are on a regular grid over the input region, as shown in Fig.~\ref{fig:schematic}(b).
We keep the RBF centers fixed and tune the amplitudes $a_{ij} \in \mathbb{R}$.
The radial offset $\Phi(\bs{s}^\prime)$ at a given normalized cone point $\bs{s}^\prime$ is defined as the output of the RBF network with Gaussian kernels, an example of which is shown in Fig.~\ref{fig:rbfsample}:
\begin{equation}
  \Phi(\bs{s}^\prime) = \sum_{i,j=1}^{N} a_{ij} \mathrm{RBF}(\bs{s}_{ij},\bs{s}^\prime),
  \quad \text{where}\
  \mathrm{RBF}( \bs{s} , \bs{s}^\prime) =
  \exp\left( -\frac{\|\bs{s} - \bs{s}^\prime\|^2}{2 \beta} \right)
\end{equation}
When computing the Cartesian coordinates for a point on the cone, parameterized by a height $s_1$ and an angle $s_2$, first we compute the RBF offset $\Phi(\bs{s^\prime})$, and then add this offset to the radius.
The surface normals of the outer cone can be computed as the cross product of the two partial derivatives of the Cartesian coordinates w.r.t. the parameters.
This cross product also has a dependence on the amplitudes associated with the RBF centers, which will be used as the model parameters during minimization.
Changing the RBF amplitudes causes a change in the surface normals, which in turn changes the direction of the refracted light rays, and by a consequence the direction and length of the distortion vectors.

\section{Raycasting Model}\label{sec:raycast}

\begin{figure}[t]
	\centering
\begin{tikzpicture}[every node/.style={on grid,draw,minimum size=2.5em,circle},node distance=14mm and 18mm,outer sep=0pt, inner sep=0pt, >=latex,thick]
	\node (p) {$\bs{p}$};
	\node[above right=of p] (rcam) {$\bs{r}_{cam}$};
	\node[below=of rcam] (xi) {$\bs{x}_i$};
	\node[below=of xi] (ni) {$\bs{n}_i$};
 	\node[right=of rcam] (rm) {$\bs{r}_m$};
	\node[below=of rm] (xo) {$\bs{x}_o$};
	\node[below=of xo,blue!60!black] (no) {$\bs{n}_o$};
	\node[above right=0.2cm of no.east,draw=none,blue!60!black,rectangle,anchor=west] {\footnotesize\begin{tabular}{ll}Function of\\ RBF network\end{tabular}};
	\node[right=of rm] (ro) {$\bs{r}_o$};
	\node[below=of ro] (xt) {$\bs{x}_t$};
	\node[right=of xt] (xcb) {$\bs{x}_{cb}$};
	\foreach \einit/\efinal in %
	  { p/rcam, rcam/xi, xi/ni, rcam/rm, ni/rm, xi/xo, rm/xo,%
	    xo/no, rm/ro, no/ro, xo/xt, ro/xt, xt/xcb%
	  }
	  \draw[->] (\einit)--(\efinal);
\end{tikzpicture}
	\caption{The computation graph of raycasting: given a pixel $\bs{p}$, we compute the 3D point $\bs{x}_t$ on the checkerboard plane, respectively the local coordinate $\bs{x}_{cb}$ of the corners. Notation: $\bs{r}$ -- ray direction, $\bs{x}$ -- Cartesian coordinate vector, and $\bs{n}$ -- normal vectors.}
	\label{fig:depgraph}
\end{figure}
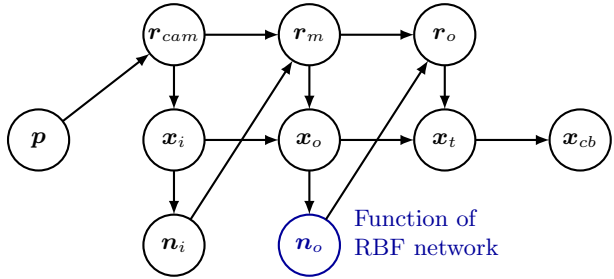

The ray casting model describes how we associate a pixel from the image with a 3D point on an object, in our case on a checkerboard pattern. In a distortion-free setup this can be achieved using the pinhole camera model, which uses a perspective projection and a set of linear operations to describe this relationship. In the presence of a refractive surface this simple perspective geometric description does not hold, and we need additional steps to associate pixels with world points. The complete computational graph can be seen on Fig.~\ref{fig:depgraph}.
Each ray starts at the camera center and we assume that the intrinsic camera parameters -- the focal lengths and the principal point, ie. the intrinsic camera matrix -- are known. All 3D points are expressed in the camera coordinate system. 
Using the camera intrinsics, we can convert any pixel coordinate $\bs{p}$ to metric coordinates, which after normalization correspond to the direction vector $\bs{r_{cam}}$ of the light ray that passes through the selected pixel.

The light ray coming from the camera first hits the inner side of the refractive surface.
Using the fact that the inner surface is a regular cone, we first compute the intersection with the cone $\bs{x}_i$ and its normal $\bs{n}_i$.
The direction $\bs{r}_m$ of the refracted light-ray inside the media is computed using Snell's law \cite{pharr2016physically}.
Knowing the geometry of the refractive body and the new direction of the refracted ray, we first identify the location $\bs{x}_o$ where the ray hits the outer surface, and is refracted for the second time. Using the surface normal $\bs{n}_o$ at this point we compute the direction of the outgoing light-ray, which we denote with $\bs{r}_o$.
Note that this second refraction is modulated by the \emph{direction} of the normal, that is parameterized by the RBF network.
At the same time, we ignore the changes caused by the RBF network in the \emph{thickness} of the material when computing $\bs{x}_o$ and we argue that this approximation holds, as the offsets are significantly smaller than the distance between the two cones (the thickness of the media), and the difference is negligible.

Finally, this outgoing light ray hits the calibration target -- in our case the checkerboard pattern -- whose position is defined through a 3D rotation and the 3D translation of the board center relative to the camera coordinate system.
The intersection point $\bs{x}_t$ is computed as an intersection of a line and a plane. For an easier handling of the checkerboard, we define a local 2D coordinate system on the object plane, with its origin at the board center, and the two axes being the horizontal and vertical directions of the square grid.
We denote the local coordinates of a 3D point $\bs{x}_t$ as $\bs{x}_{cb}$.

\section{Optimization of the Surface Parameters}\label{sec:optim}

The estimation of image distortions is equivalent to finding the surface parameters that generated a set of calibration images.
We use a square checkerboard pattern as calibration target, and take multiple images of the same target in different positions.
We use gradient descent minimization to find the amplitudes $\bs{a} = \left\{a_{ij}\right\}$, $i,j=\overline{1, N}$, of the RBF centers, the parameters of the refractive surface.
During the minimization all other parameters, including the camera intrinsics, the sizes of the inner and outer cones, as well as the calibration pattern pose and size are assumed to be known.

For each calibration image $\bs{I}^k$, $k = \overline{1, N_i}$ we find the pixel coordinates and ordering of the checkerboard pattern corners $\left\{\bs{p}_{ij}^k\right\}$, $i,j=\overline{1, N_c}$.
The corresponding local coordinates of the detected corners on the object plane -- denoted with $\left\{\bs{x}_{ij}^{cb}\right\}$, $i,j=\overline{1, N_c}$ -- are given by their distance from the board center.
Let $f_{\bs{a},k}(\cdot)$ be the raycasting function described in Sect.~\ref{sec:raycast}, parameterized by the RBF amplitudes $\bs{a}$, which takes an input pixel, and computes the local coordinates of the corresponding world point on the $k-th$ target checkerboard pattern.
Then the loss function used for the minimization is the ${\cal{L}}_2$ loss between the local coordinates of a corner estimated by the raycast function, and the ground-truth local coordinates of the corners:
\begin{equation}
	\mathcal{L}(\bs{a}) = \sum_{k=1}^{N_i} \sum_{i,j = 1}^{N_c} \left\lVert f_{\bs{a}, k} \left(\bs{p}_{ij}^k \right) - \bs{x}_{ij}^{cb} \right\rVert^2
\end{equation}
The optimal parameters $\bs{a}^\star$ are found through a Gradient Descent minimization of the loss function:
\begin{equation}
	\bs{a}^\star = \arg\min_{\bs{a}} \mathcal{L}(\bs{a})
\end{equation}
The optimization algorithm including the ray-casting model is implemented using the PyTorch \cite{paszke2017automatic} deep learning framework.
This framework makes the implementation easier as it provides backward automatic differentiation and implements gradient descent minimization.

\begin{figure}[t]
	\centering
	\begin{tabular}{@{}p{0.35\linewidth}p{0.1\linewidth}p{0.35\linewidth}@{}}
		\pgfimage[width=\linewidth]{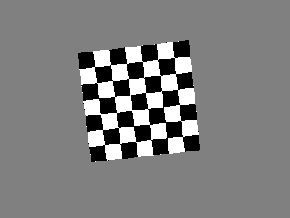}
		& &
		\pgfimage[width=\linewidth]{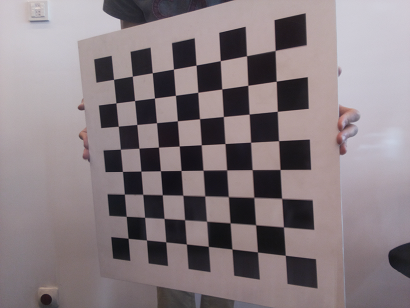}
		\\
		\centerline{(a)} & & \centerline{(b)}
	\end{tabular}
	\vspace*{-1.5em}
	\caption{Samples of (a) rendered and (b) real images used in our experiments.}
	\label{fig:boards}
\end{figure}

\section{Experiments}\label{sec:exp}

We evaluate our algorithm on two data sets: a noise-free synthetic one and a real experimental setup.
In the synthetic case we show that our algorithm is capable of finding the optimal parameters that generated a given image even for large irregularities on the outer surface.
In the second case we present an experimental setup, and show that the algorithm is able to reduce reconstruction errors in real-world scenarios.

\begin{figure}[t]
	\centering
	\begin{tabular}{@{}p{0.45\linewidth}p{0.1\linewidth}p{0.45\linewidth}@{}}
		\centerline{\pgfimage[width=\linewidth]{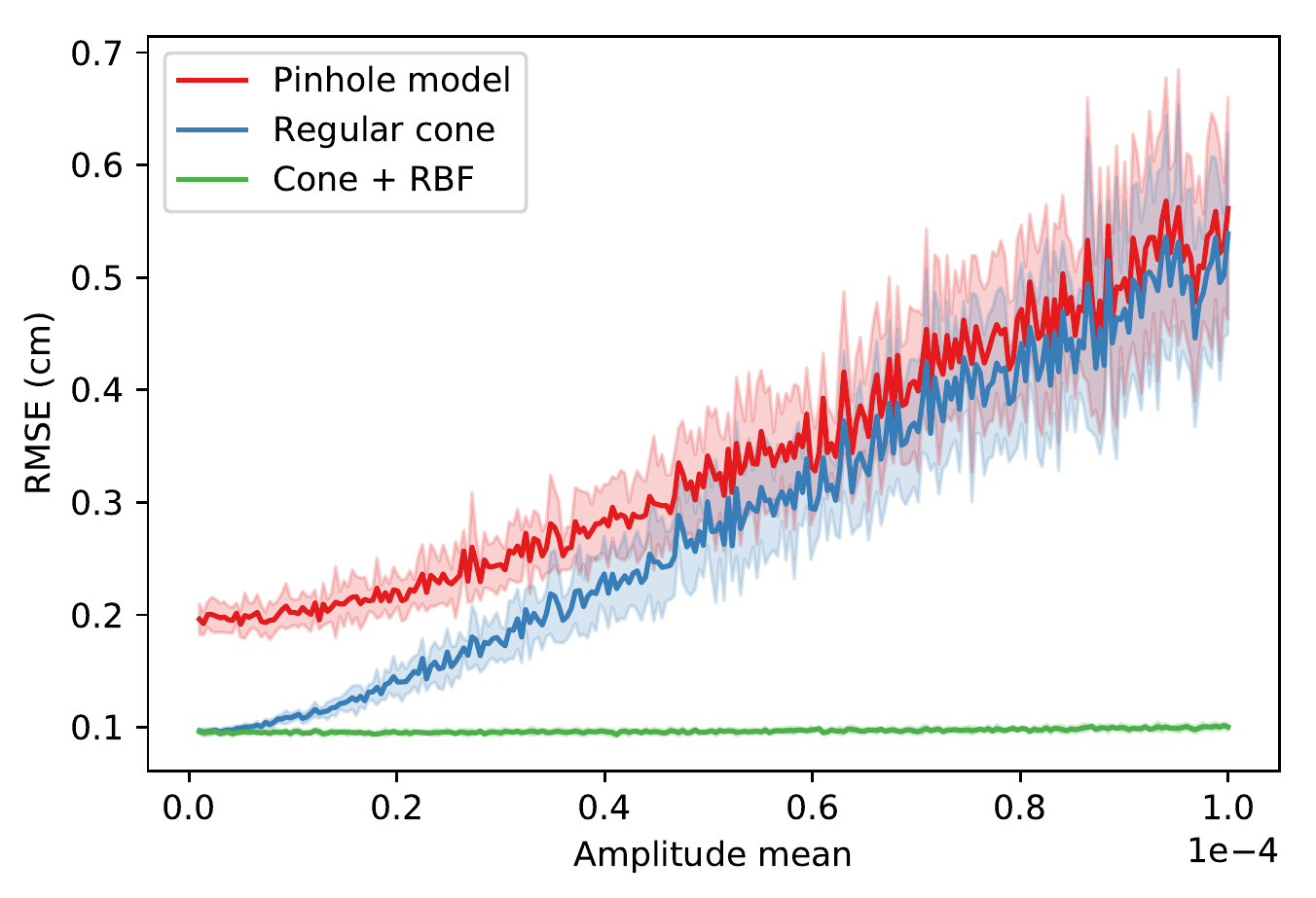}}
		& &
		\centerline{\pgfimage[width=\linewidth]{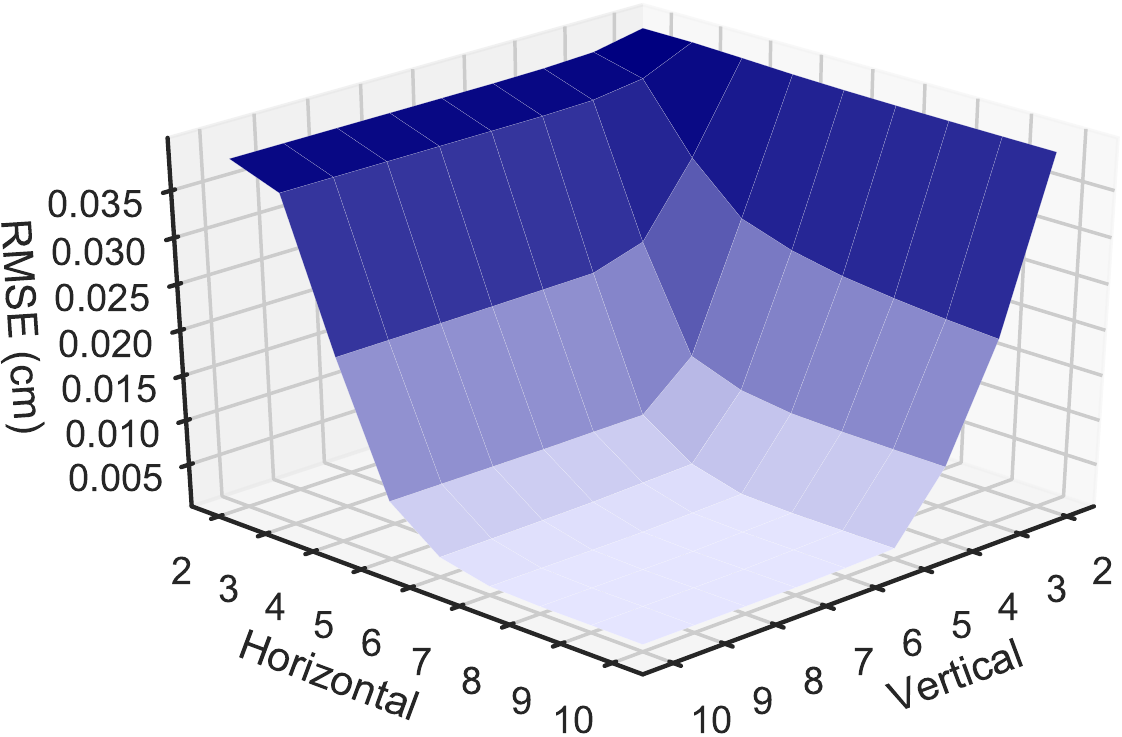}}
		\\
		\centerline{(a)} & & \centerline{(b)}
	\end{tabular}
	\vspace*{-1.5em}
	\caption{%
	Results for synthetic data.
	(a) %
	Errors for different refraction complexity: \emph{Pinhole model} disregards distortions, \emph{Regular cone} uses a perfect cone, and \emph{Cone + RBF} uses optimization.
	(b)
	 Errors for varying RBF complexities: as the number of RBF centers increases, the estimation error drops.
	}
		\label{fig:err_synth_ncenter}
\end{figure}

\subsection{Synthetic Dataset}\label{sec:exp:synth}

In a first set of experiments we applied the forward image generation model to render synthetic images.
We set the parameters of the camera, refractive surface, and the checkerboard pattern to similar values as in the real-world experiment, as shown in Fig.~\ref{fig:boards}(a).

In our first synthetic experiment, using an RBF grid of size $4\times4$, we sample the amplitudes $\bs{a}$ from a Gaussian with mean $\mu_a \in [\num{1e-6}, \num{1e-4}]$ and standard deviation $\sigma_a = \mu_a/4$.
With the sampled surface we generate $10$ synthetic images using random positions for the calibration target -- as shown in Fig.~\ref{fig:boards}.
Using the batch of $10$ images we run the optimization from Sect.~\ref{sec:optim} for 500 steps and store the final error as the  \emph{root mean squared error} between the predicted and ground-truth checkerboard corners.

For each amplitude distribution we repeat the whole process $10$ times and show the results in Fig.~\ref{fig:err_synth_ncenter}(a): \emph{Pinhole model} is the error without any distortion model, while \emph{Regular cone} and \emph{Cone + RBF} show the results using a \emph{perfect} cone and one where parameters are inferred.
We see that with small RBF amplitudes the error is a result of the cone geometry, while with higher amplitudes the errors due to the uneven surface dominate the distortion.
An important conclusion is that for this case, where is no observation noise, the optimization algorithm works well, reducing the errors almost to zero.

In the second experiment we generate random refraction patterns using $10\times10$ RBF centers with amplitudes drawn randomly from Gaussian distributions, and show the errors after performing the optimization -- corresponding to the \emph{Cone + RBF} case from Fig.~\ref{fig:err_synth_ncenter}(a).
Instead of changing the amplitudes, we vary the numbers of RBF centers -- both horizontally and vertically, results are shown in Fig.~\ref{fig:err_synth_ncenter}(b).
We observe that -- in this artificial setup -- the refractive surface is well approximated with a smaller complexity model: $7\times5$ parameters are enough to approximate the generated surface (the asymmetry is due to our setup of the refractive surface, which is a cone with larger horizontal curvature).

\subsection{Real Dataset}

\begin{table}[t]
	\centering
	\caption{Root Mean Squared Errors for the 3 real sets of images. Initial error considers no distortion model, final error is obtained using Cone + RBF surface model with optimized parameters. Last column shows the relative improvement.}
	\begingroup
	\renewcommand{\arraystretch}{1.5}
	\begin{tabular}{l||c|c|c}
	Set & $RMSE$ initial (\si{\centi\meter}) & $RMSE$ final (\si{\centi\meter}) & Rel. imp. \\
	\hline
	1 & 0.1364 & 0.0772 & 43.35\% \\
	2 & 0.1649 & 0.0941 & 42.90\% \\
	3 & 0.1570 & 0.0973 & 38.04\% \\
	\end{tabular}
	\endgroup
	\label{tab:rmse_real}
\end{table}

For a real-world experiment we use a Raspberry Pi Camera Module v2 to capture the checkerboard images.
The camera has a $3.68 \times 2.76$ \si{\milli\meter} sensor and registers images on a $3280 \times 2464$ pixel resolution.
Prior to the experiment we calibrated the camera using Zhang's method \cite{zhang2000flexible} and we registered a $2558.36$ pixel focal length and a principal point at the $(1666.03, 1273.65)$ location.
After calibration we placed a cone shaped glass in the front of the camera, with approximate parameters shown on Fig.~\ref{fig:schematic}(b).
In our experiments we use $45$ images of a checkerboard pattern that were randomly split into $3$ non-overlapping sets of $15$ items each, which we will refer to as Sets 1, 2 and 3.

Since the positions of the checkerboard patterns are unknown, we have to estimate them.
We do this in two steps: (1) we estimate the object pose by running Zhang's method on the distorted images, while we fix the camera intrinsic parameters to the calibrated values; and (2) we apply the \emph{perfect} cone refraction model and with the same objective function we minimize for the calibration pattern positions.
Keeping the above values fixed, we then run the optimizer for the RBF center amplitudes.

\begin{figure}[t]
	\begin{tabular}{@{}p{0.48\linewidth}p{0.02\linewidth}p{0.48\linewidth}@{}}
		\centerline{\pgfimage[width=\linewidth]{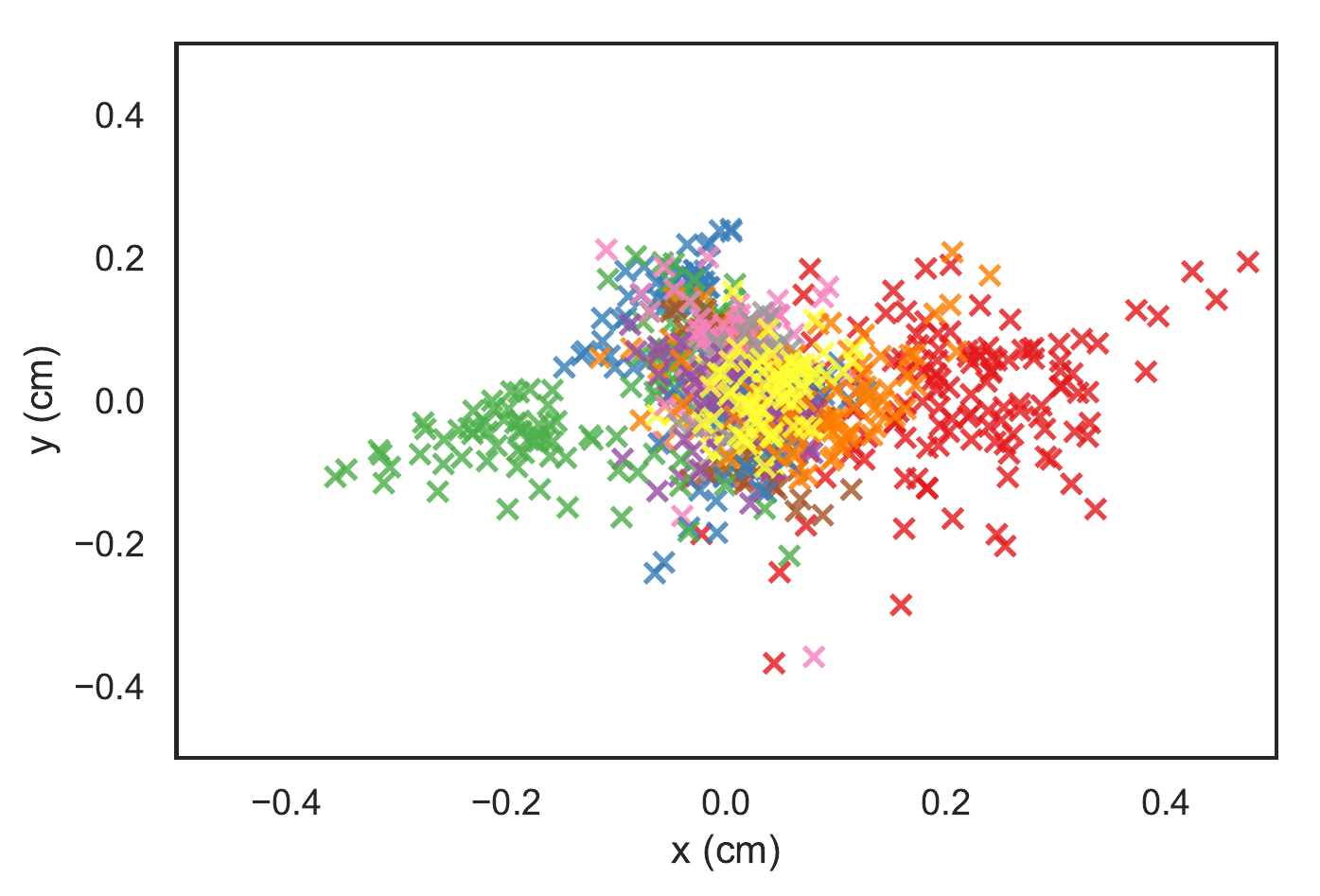}}
		& {} &
		\centerline{\pgfimage[width=\linewidth]{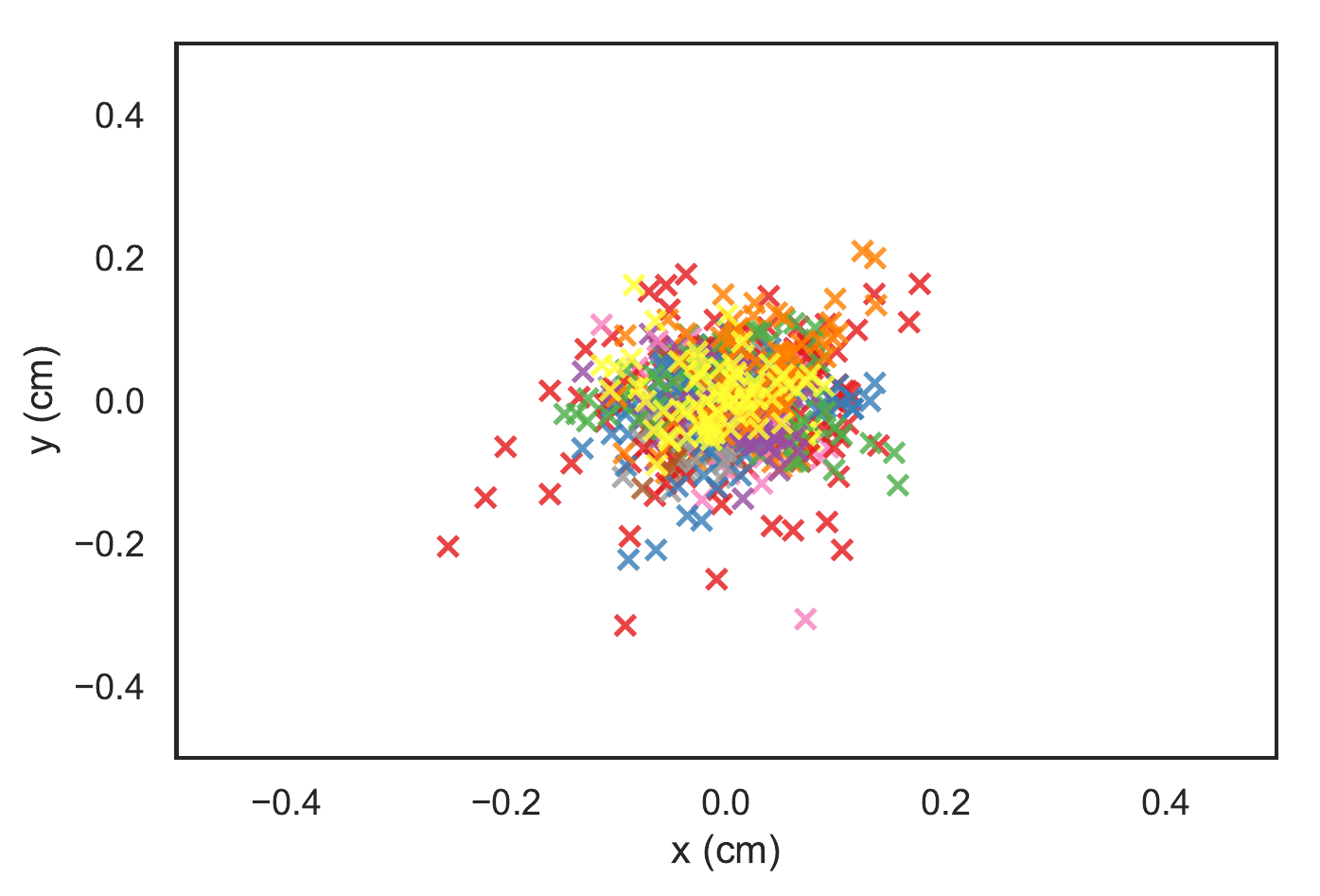}}
		\\
		\vspace*{-2em}\centerline{(a)} & {\vspace*{-2em}} & \vspace*{-2em}\centerline{(b)}
	\end{tabular}
	\vspace*{-2em}
	\caption{Scatter plots of all the checkerboard corner errors. $x$ and $y$ are the axis in the local coordinate system of the checkerboard patterns (different for each image). Different colors correspond to different images when (a) no distortion model is used, and when (b) the optimized distortion model is used.}
	\label{fig:err_scatter}
\end{figure}

With our real dataset we use $8 \times 8$ RBF centers -- found using experimentation as in Fig.~\ref{fig:err_synth_ncenter} -- and run the minimization for $1000$ steps and we report results for the different sets separately.
Table \ref{tab:rmse_real} shows the RMSE for the 3 sets of images.
The errors can be interpreted as the 3D distance in centimeters between the ground-truth and the predicted position of the same corner on the checkerboard pattern.
The first column shows the mean error in the case where we consider no distortion model at all, while the second column shows the results after the minimization using the \emph{Cone + RBF} surface model.
Last column shows the relative improvements between the two error values.
Our method is able to improve the 3D distances in each case, resulting in a more precise generative model of the image formation process.
Fig.~\ref{fig:err_scatter} shows all corner errors in the 2D local coordinate system of the checkerboard, with different colors for different images.
We see that in the uncorrected scatter plot (a) for each image -- denoted by the color -- there is a dominant direction, which is caused by the horizontal curvature of the cone.
We highlight that the effect of minimization -- right subplot in Fig.~\ref{fig:err_scatter} -- is that the errors are around zero, without dominant directions for individual images.

\subsection{Analyzing Distortions}

Most of the distortion estimation methods directly model the pixel displacements on the image plane, and define a single, fixed distortion map for a given camera.
In contrast, our model estimates the distortion map by explicitly modeling the refractive material and using a ray-tracing.
We consider this advantageous since the -- usually separate directional distortions -- are given a unified and \emph{consistent} generative model.
A consequence is that the physical model introduces a depth-dependent component in the distortion map, where the distance of a 3D point has to be known in order to find the image distortion, that is usually expressed in pixels.

In order to compute the image distortion vector, we start from a distorted pixel on the image.
Using the distorted pixel $\bs{p}_d$, we use the ray-casting algorithm to obtain a 3D coordinate $\bs{x}_t$ of the object point at a given distance.
The undistorted pixel coordinates $\bs{p}_u$ for an object point can be computed using the pinhole camera model.
The distortion vector $\Delta\bs{p}$ for a given pixel $\bs{p}_d$ is given by the difference between the distorted and undistorted coordinates, i.e.\ $\Delta\bs{p} = \bs{p}_u - \bs{p}_d$.

\begin{figure}[t]
	\centering
	\pgfimage[height=5cm]{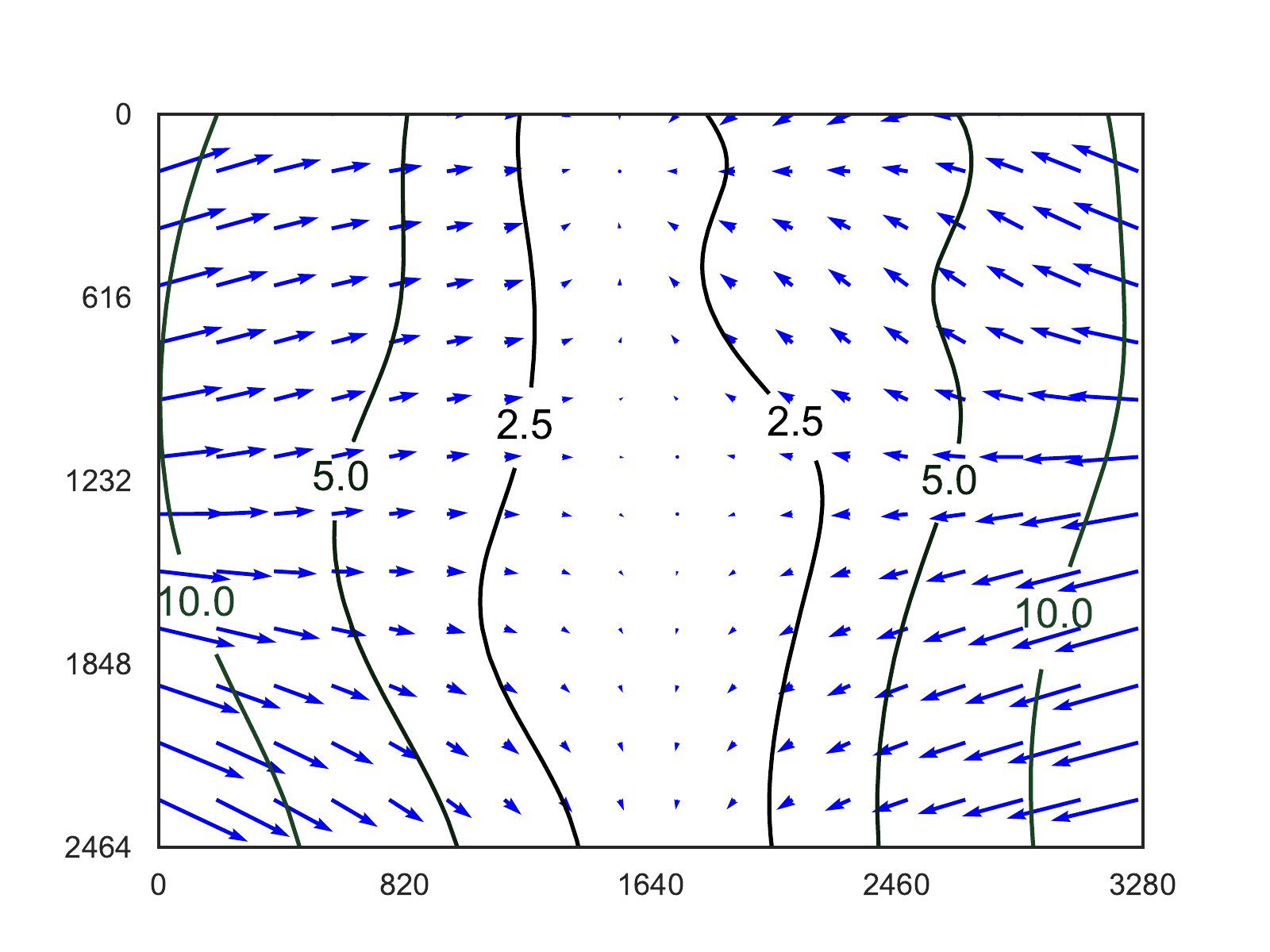}
	\caption{Distortion vectors for Set 1 for a fixed object distance of \SI{1}{\meter} with contour lines for distortions of $2.5$, $5$, and $10$ pixels. The curvature of the cone shaped glass object introduces large horizontal distortions on the two lateral image sides.}
	\label{fig:dist_field}
\end{figure}

Fig.~\ref{fig:dist_field} shows the image distortion vector field using the optimal parameters for the images from Set 1.
To better visualize our model and to be able to compute a distortion field, we fixed the object distance to \SI{1}{\meter} for the whole scene; this is comparable to the range of the calibration objects on the images.
Rays corresponding to the middle pixels are almost orthogonal to the refractive surface, resulting in little or no distortion, while the rays located on the two lateral sides have a large angle compared to the surface normal, resulting in large refraction and large image distortions.

\begin{figure}[t]
	\centering
	\pgfimage[height=5cm]{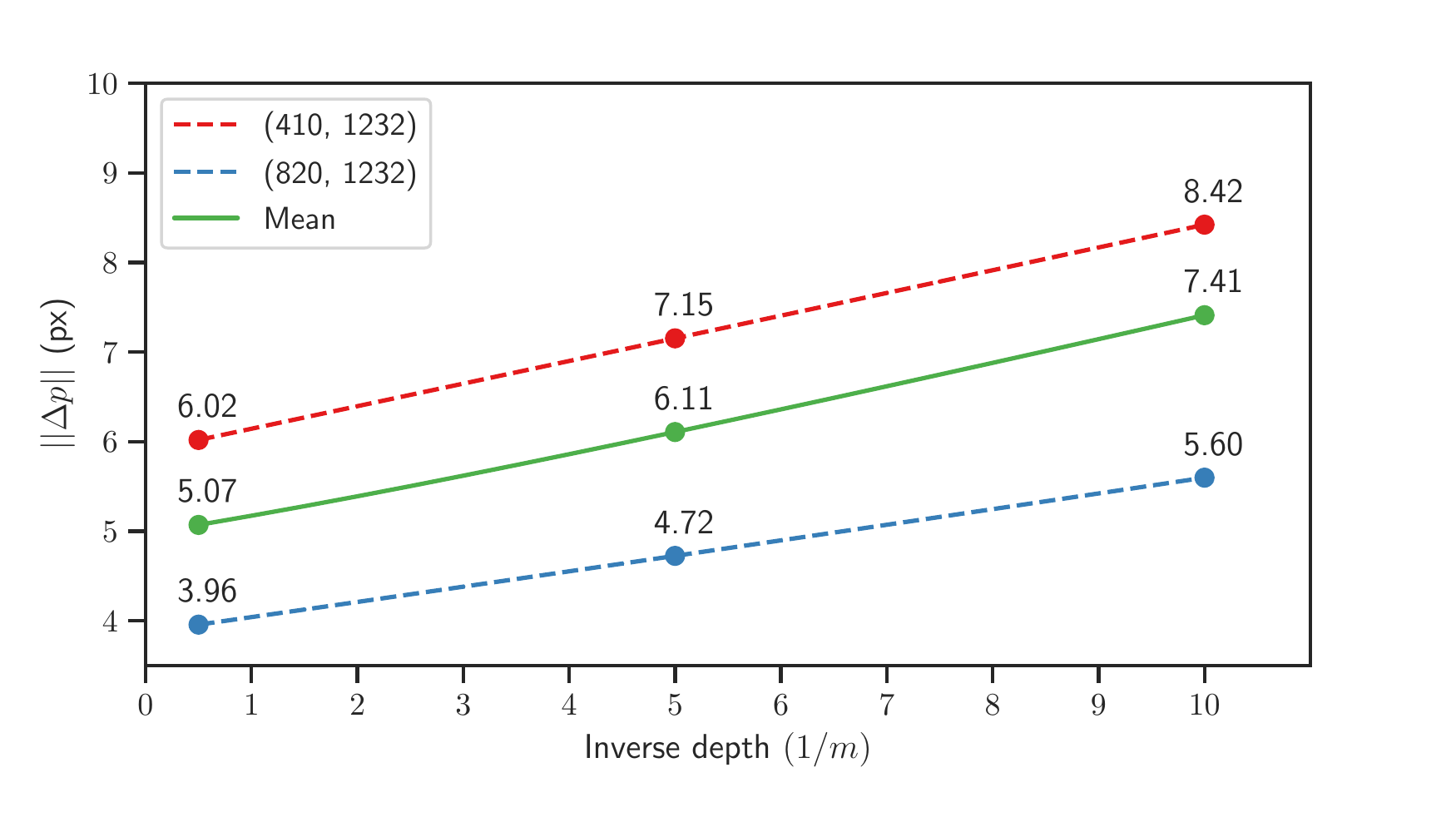}
	\caption{Individual and average pixel distortions as a function of the \emph{inverse depth}. We see a linear dependence between the inverse depth and the distortion norm.}
	\label{fig:dist_depth}
\end{figure}

A consequence of the physical \emph{generative} model is the possibility to inspect the dependence of the distortions on pixel depth, as shown in Fig.~\ref{fig:dist_depth}.
The image shows the change of distortion norm for two individual pixels with coordinates $(820, 1232)$ and $(410, 1232)$, as well as the change of the mean distortion norm.
We can observe, that the distortion norm shows a linear dependence on the inverse depth of the pixel.
The figure shows the distortions for the inverse depth interval of $0.5$ (corresponding to a depth of \SI{2}{\meter}) to $10$ (depth of \SI{10}{\centi\meter}).
We can also observe that the slope of the line showing the change of distortion increases as we get closer to the edges of the image, where the change of distortion is $2.4$ pixels over the analyzed interval, compared to the change of $1.64$ pixels for the point closer to the image center.

\section{Conclusions}

In our work we presented a model for geometric image distortions caused by refractive surfaces being placed between the camera and the scene.
Based on an explicit model of the refractive surface, we presented a forward generative model of the distortions and the image generation; the generating process used the ray-casting mechanism.
We assumed a conic refractive surface and used an additive model for the imperfections of the surface; the used model was a restricted Radial Basis Function network.
Using model inversion and automated differentiation, we estimated the refractive surface with a set of checkerboard calibration target  images.
We validated our algorithm on synthetic and real-world data, and analyzed the observed image distortions.

The benefit of the method is that the model we proposed \emph{is parametric}: despite being strongly non-linear and complex, using a small set of calibration images, the algorithm is able to find the global distortion map.
A second important aspect is that the data collection  methodology allows for heterogeneous data: by estimating the view angles, we can use the \emph{whole} dataset for estimation.

A weak limitation of our method is the fixed base shape, which in the current formulation was a cone.
We chose this shape as it was the closest to the actual object used in the experiments.
A work-around to the strict constraint of the shape is to parameterize it and -- added to the parameters of the irregularities -- to optimize for an extended set of parameters.
Evidently, as the number of free parameters grows, this introduces further modeling difficulties, as the degree of freedom increases and the optimization can be much harder.

Future work will focus on (1) developing a model that is general enough to be suitable for a wide range of applications without significant changes, and (2) will explore whether there is a possibility for our method to infer distortions using other types of inputs to the system.
An extension possibility is to use -- under specific constraints -- other data, like the estimated optical flows with certain constraints on the collection procedure.

\bibliographystyle{splncs04}

\end{document}